\documentclass{elsarticle}

\pdfoutput=1 
\usepackage[T1]{fontenc}    
\usepackage[utf8]{inputenc} 

\usepackage{graphicx} 
\DeclareGraphicsExtensions{.pdf,.png,.eps,.ps}
\graphicspath{{./figures/}}

\usepackage{amsmath,amsthm, amssymb} 
\usepackage{enumerate}
\usepackage{float}
\usepackage[Algorithm]{algorithm}
\usepackage{algpseudocode}
\usepackage{subfigure}
\usepackage[hidelinks]{hyperref}

\newtheorem{theorem}{Theorem}
\newtheorem{proposition}{Proposition}
\newtheorem{lemma}{Lemma}
\newtheorem{definition}{Definition}

\newtheorem{property}{Property}

\theoremstyle{definition}
\newtheorem{example}{Example}

\newcommand{\RR}{{\mathbb R}}
\newcommand{\phifreeG}{\paths(\chifree_G)}

\newcommand{\graphs}{\mathcal{G}}
\newcommand{\paths}{\Phi}
\newcommand{\NN}{{\mathbb N}}
\newcommand{\robots}{\mathcal{R}}
\newcommand{\phifree}{\paths(\chifree)}
\renewcommand{\path}{\varphi}

\newcommand{\chifree}{\chi^{\mathrm{free}}}
\newcommand{\chiobs}{\chi^{\mathrm{obs}}}

\newcommand{\vmax}{\overline{v}}
\newcommand{\uLbf}{\underline{\mathbf{u}}}
\newcommand{\uL}{\underline{u}}
\newcommand{\uHbf}{\overline{\mathbf{u}}}
\newcommand{\uH}{\overline{u}}
\newcommand{\controls}{\mathbf{U}}
\newcommand{\ubf}{\textbf{u}}
\newcommand{\dt}{\Delta T}
\newcommand{\e}{\mathbf{e}}

\newcommand{\imp}{\mathrm{impulse}}
\newcommand{\im}[1]{{\mathrm{Im}\left(#1\right)}}

\usepackage[usenames,dvipsnames]{color}

\journal{Systems and Control Letters}

\begin{document}

\begin{frontmatter}



\title{Robust multirobot coordination using \\priority encoded homotopic constraints}

\author[mines]{Jean Gregoire\corref{cor1}}
\ead{jean.gregoire@mines-paristech.fr}
\author[mines]{Silvère Bonnabel}
\ead{silvere.bonnabel@mines-paristech.fr}
\author[mines,inria]{Arnaud de La Fortelle}
\ead{arnaud.de\_la\_fortelle@mines-paristech.fr}


\address[mines]{MINES ParisTech, Center for Robotics\\60, boulevard Saint Michel\\75272 PARIS Cedex 06, FRANCE\\ Phone number: +33(0)140519408
}
\address[inria]{Inria Paris - Rocquencourt, RITS team}

\cortext[cor1]{Corresponding author}

\begin{abstract}
We study the problem of coordinating multiple robots along fixed geometric paths. Our contribution is threefold. First we formalize the intuitive concept of priorities as a binary relation induced by a feasible coordination solution, without excluding the case of robots following each other on the same geometric path. Then we prove that two paths in the coordination space are continuously deformable into each other if and only if they induce the \emph{same priority graph}, that is, the priority graph uniquely encodes homotopy classes of coordination solutions. Finally, we give a simple control law allowing to safely navigate into homotopy classes \emph{under kinodynamic constraints} even in the presence of unexpected events, such as a sudden robot deceleration without notice. It appears the freedom within homotopy classes allows to much deviate from any pre-planned trajectory without ever colliding nor having to re-plan the assigned priorities.
\end{abstract}


\end{frontmatter}


\section{Introduction}

We consider the problem of coordinating a collection of cooperative robots at an intersection area along fixed geometric paths, motivated by applications such as coordinating a fleet of automated guided vehicles in a factory, or automated cooperative vehicles in a fully automated transportation system. This coordination problem has been extensively studied and is formulated in the so-called coordination space, first introduced in~\cite{ODonnell1989} and become standard~\cite{LaValle2006}. Due to the promises in autonomous cars design, automated intersection management has attracted much interest recently~\cite{Dresner2004, Colombo2012, Kim2014}, but the focus is primarily on the efficiency of the trajectory planner and of the trajectory tracking~\cite{Falcone2007}. The complexity of searching a time-optimal optimal trajectory in the coordination space grows however exponentially with the number of robots involved~\cite{Hopcroft1984}, and all the proposed solutions are thus necessarily based on heuristics to allow for real-time effective trajectory computation. In contrast, References~\cite{Lumelsky1987,Pallottino2007} proposed coordination without planning at all, based on simple interconnected primitive reactive behaviors~\cite{Brooks1986}. It allows for reactivity and robustness to uncertainty, but as noticed in~\cite{Lumelsky1987}, efficiency or even deadlock avoidance guarantees are difficult to obtain. 

This paper proposes to use homotopy constraints in multirobot coordination to combine the benefits of reactivity -- handle unexpected events in a reactive manner -- and of deliberation -- in particular, deadlock avoidance guarantee. Deliberation consists of choosing a particular homotopy class and the control part boils down to navigation in the assigned homotopy class using the freedom of action within the homotopy class to allow for reactivity. Previous work already noticed the advantages of planning a set of homotopic paths to allow for more reactivity than single path planning, in particular using the concept of elastic strips (see, e.g.,~\cite{Quinlan1993,Brock2002}). It is also possible to obtain optimality results, by searching for an optimal path within the assigned set of homotopic paths~\cite{Bhattacharya2013}. In multirobot coordination, this approach based on homotopy considerations has been poorly used. In~\cite{Ghrist2005,Ghrist2006}, the existence of homotopy class of feasible paths in the coordination space is noticed. The authors provide a way to deform a given path into a Pareto-optimal path of the homotopy class, focusing on trajectory planning and the efficiency of the planned trajectories. In this paper, we propose to focus on robustness aspects, aiming to allow for reactivity with respect to unexpected events, using homotopy considerations in the coordination space. 

The contribution of the paper is twofold. First of all, we go beyond the sole proof of existence of homotopy classes in the coordination problem noticed in~\cite{Ghrist2005,Ghrist2006} by providing a meaningful unique representative of homotopy classes: the priority graph. Feasible priority graphs are in bijection with homotopy classes of solutions to the coordination problem as stated in Theorem~\ref{thm:invariance-priority-graph}. The second contribution is a control law ensuring navigation within some assigned homotopy class, given in the form of a priority graph (see Theorem~\ref{thm-robustness-brake-application}). The benefits of using homotopic constraints come from the freedom of action available within an assigned homotopy class allowing for more reactivity than under the execution of a planned trajectory. In particular, events requiring deceleration of some or all robots can be handled in a reactive manner without changing priorities. This may be particularly useful in an autonomous driving context where vehicles and pedestrians share the road resulting in a particularly unpredictable environment, with many events requiring momentary deceleration or even stop. 

The paper is organized as follows. Section~\ref{sec:priorities} formalizes the intuitive concept of priorities for the coordination of multiple mobile robots along fixed geometric paths. Section~\ref{sec:priority-encoded-homotopy-classes} proves the existence of homotopy classes of solutions to the multirobot coordination problem, uniquely encoded by priority graphs. Section~\ref{sec:control} focuses on the control of robots under assigned -- priority encoded -- homotopic constraints and our approach's robustness is illustrated in Section~\ref{sec:robustness} including simulations. Section~\ref{sec-conclusion} concludes the paper and opens perspectives for future work.

\section{A novel tool: the priority graph}
\label{sec:priorities}
\subsection{The coordination space approach}

\begin{figure}[!htbp]
\begin{center}
\includegraphics[width=0.7\linewidth]{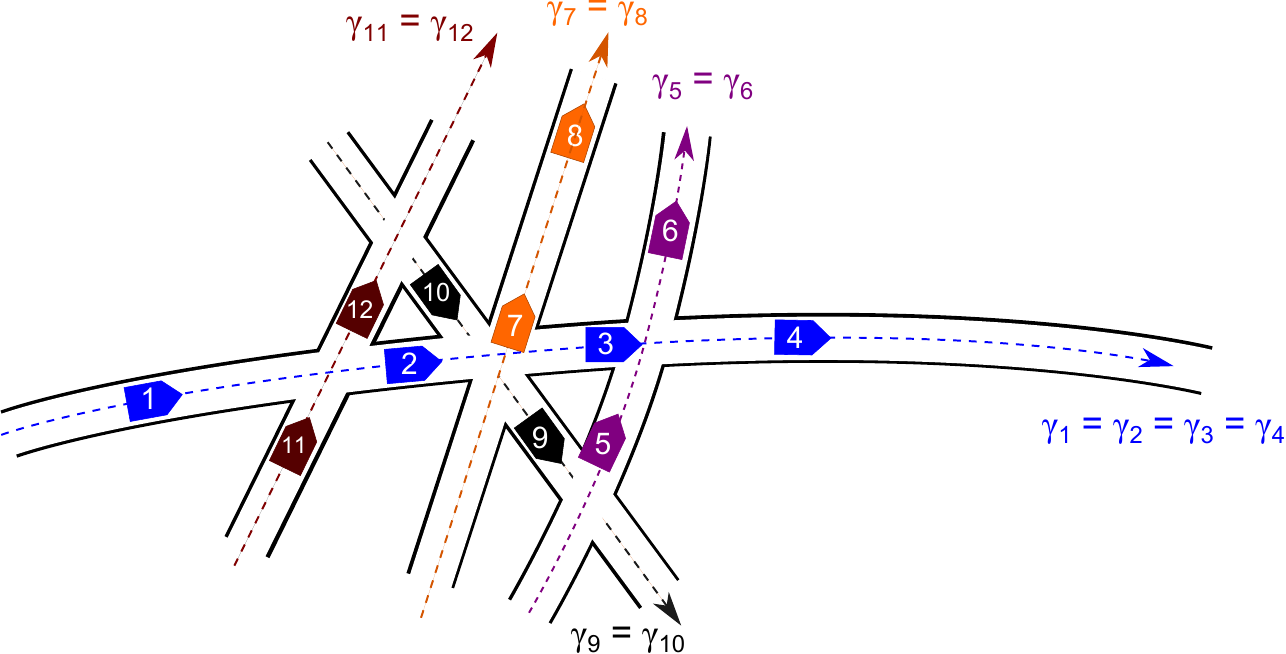}\hfill
\end{center}
\caption{The fixed paths assumption. Every robot travels along an assigned path.}
\label{fig-paths}
\end{figure}

Consider the problem of coordinating the motion of a collection of robots $\robots$ in a two-dimensional space. Every robot $i\in\robots$ follows a particular path $\gamma_i: x_i\in[0,1] \mapsto \gamma_i(x_i)\in\RR^2$ and we let $x_i \in [0,1]$ denote the position of robot $i$ along path $\gamma_i$ (see Figure~\ref{fig-paths}).
$x:=(x_i)_{i\in\robots}$ indicates the configuration of all robots; $x\in \chi:=[0,1]^\robots$. The configuration space $\chi$ is known as the coordination space, first introduced in~\cite{ODonnell1989} and which has become a standard tool~\cite{LaValle2006}. This approach is often referred as path-velocity decomposition. It reduces the problem's complexity as each robot has now only one degree of freedom. For an application to autonomous driving, this additional constraint seems particularly well adapted as the road network is strongly spatially organized (roads and lanes with markings). In the rest of the paper,  $\{\e_i\}_{i\in\robots}$ denotes the canonical basis of $\chi$.

Some configurations must be excluded to avoid collisions between robots (see Examples~\ref{ex:collision1} and~\ref{ex:collision2}). The obstacle region $\chiobs\subset\chi$ is the open set of all collision configurations. Let $\kappa_{ij}\subset [0,1]^2$ denote the set of couples of positions $(x_i,x_j)$ where $i$ and $j$ collide. Let $\chiobs_{ij} \subset\chi$ denote the set of configurations $x$ where $i$ and $j$ collide, we have:
\begin{equation}
\chiobs_{ij}:=\left\{x\in\chi: (x_i,x_j)\in\kappa_{ij}\right\}
\end{equation}
We obviously take $\chiobs_{ii}:=\emptyset$.
\begin{definition}[Obstacle region, Obstacle-free region]
The obstacle region is the set $\chiobs\subset\chi$ of configurations where a collision occurs for some $i,j\in\robots$, i.e., 
\begin{equation}
\chiobs := \cup_{\{i,j\}} \chiobs_{ij}
\end{equation}
$\chifree:=\chi\setminus \chiobs$ denotes the obstacle-free space.
\end{definition}
By construction, $\chiobs_{ij}$ is a cylinder (based on the plane generated by $\e_i$ and $\e_j$), and the obstacle region merely appears as the union of $n(n-1)/2$ cylinders~\cite{LaValle2006} corresponding to as many collision pairs. Every cylinder $\chiobs_{ij}$ is assumed to have an open convex cross-section, i.e., $\kappa_{ij}$ is open. 

Finally, we assume that that positions $0$ and $1$ are safe for all robots, i.e., $\chiobs\subset(0,1)^\robots$. It is rather technical and models the fact that coordination is only considered within a bounded area, collision avoidance before and after the intersection being ensured by another coordination system and not considered here.
\vspace{-1.2cm}
\begin{figure}[!htpb]
\begin{center}
\raisebox{-0.5\height}{\includegraphics[width=0.35\linewidth]{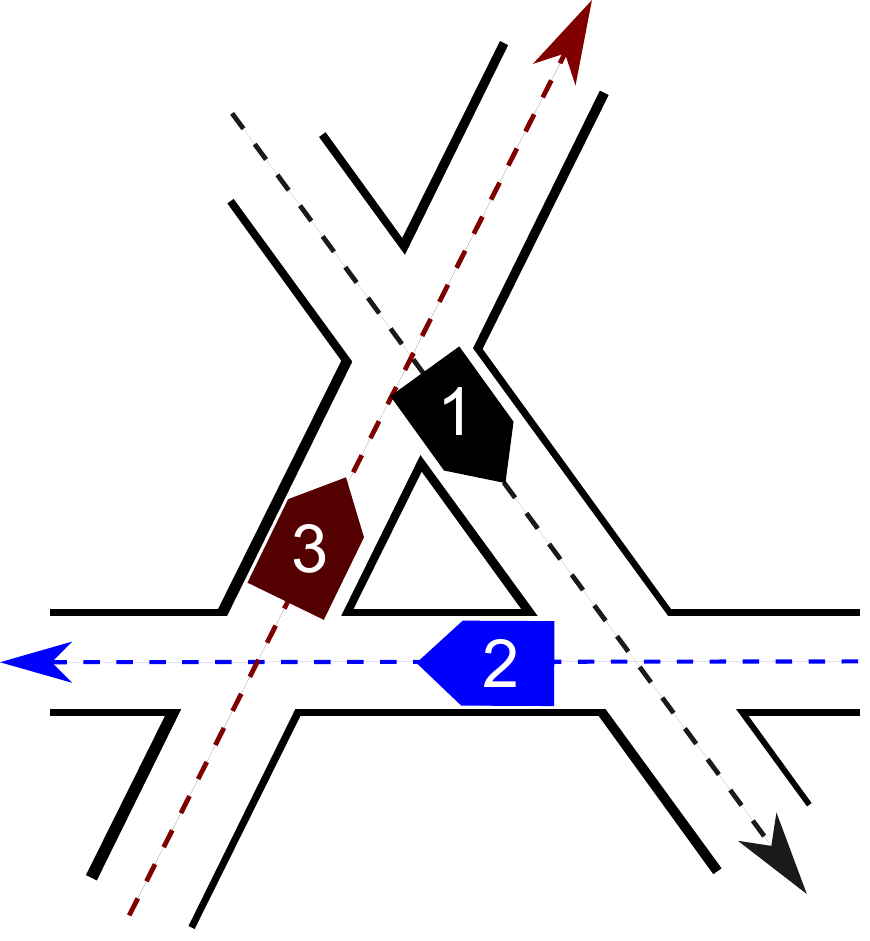}}\hfill
\raisebox{-0.5\height}{\includegraphics[width=0.65\linewidth]{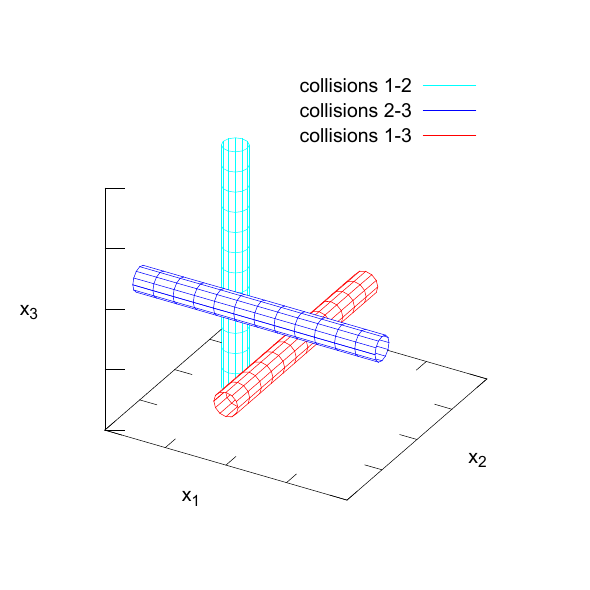}}\hfill
\vspace{-1.2cm}
\end{center}
\caption{The right drawing shows the cylindrical structure of the obstacle region for the three-robot system of the left drawing. Each cylinder accounts for the possible collisions between each couple of robots.}
\label{fig:collision-region-3D}
\end{figure}
A continuous application $\path:[0,1]\to\chi$ will be called a path and we let $\im{\path}$ denote the set of values taken by $\path$:
\begin{equation}
\im{\path}:=\left\{\path(t):t\in[0,1]\right\}
\end{equation}
A partial order $\leq$ for configurations is defined as the product order of $\RR^\robots$:
\begin{equation}
\forall x,y\in\chi, x\leq y\text{ if } \forall i\in\robots, x_i \leq y_i\\
\end{equation}

\begin{definition}[Feasible path]
A feasible path is a non-decreasing collision-free path $\path:[0,1]\to\chifree$ starting at $\path(0)=(0 \cdots 0)$ and ending at $\path(1)=(1 \cdots 1)$.
\end{definition}
We let $\paths(\chifree)$ denote the set of feasible paths. Note that we will only consider as feasible motions where robots never move backwards in the intersection area. It is a standard assumption as neither efficiency nor safety can be expected from robots moving backwards at an intersection area. 

More generally, given a subset $C\subset\chi$, we let $\paths(C)$ denote the set of non-decreasing paths satisfying $\im{\path}\subset C$, $\path(0) = (0 \cdots 0)$ and $\path(1)=(1 \cdots 1)$. This notation is coherent with the definition of $\paths(\chifree)$ as the set of feasible paths. 

In the following, we provide two examples where the obstacle region can be computed analytically.
\begin{figure}[h]
\begin{center}
\includegraphics[width=0.7\linewidth]{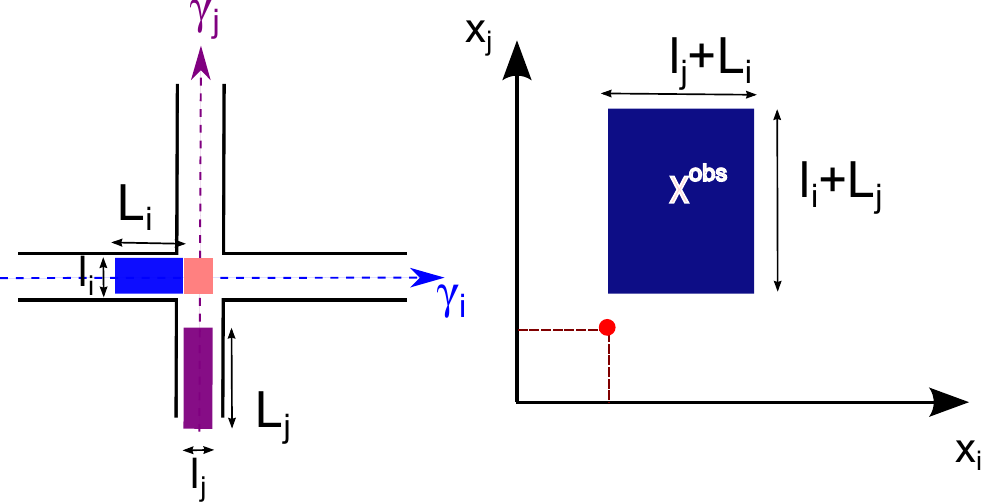}\hfill
\includegraphics[width=0.7\linewidth]{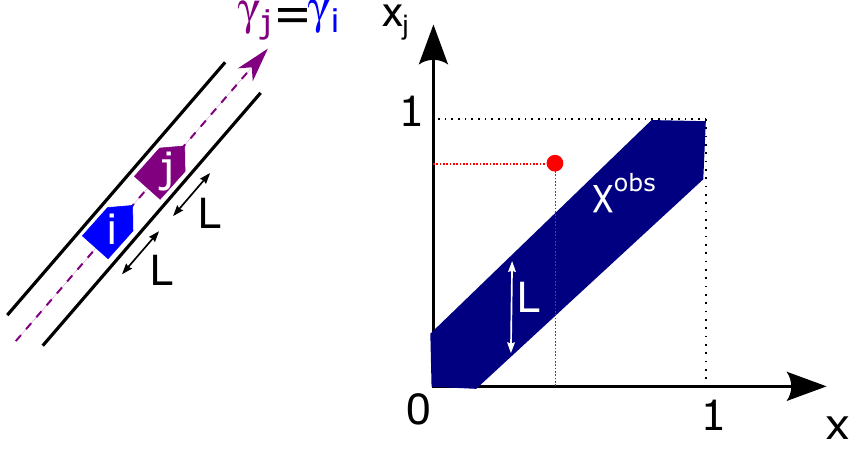}\hfill
\end{center}
\caption{Projection of the obstacle region for two rectangular robots along straight perpendicular paths (above) and for robots that follow each other (below).}
\label{fig-example-collision}
\end{figure}

\begin{example}[Two perpendicular paths with rectangular robots]
Consider two rectangular robots $i,j$ of lengths $L_i,L_j$ and widths $l_i,l_j$ along straight perpendicular paths. In the real space, there is a rectangular region of area $l_i \times l_j$ that can be occupied by only one robot, exclusively (see the red box in the top left drawing of Figure~\ref{fig-example-collision}). When a robot is at the the entry of this region (robot $i$ in the top left drawing of Figure~\ref{fig-example-collision}), it needs to travel the length of the region plus its own length in order to exit this region (robot $i$ needs to travel distance $l_j+L_i$ in order to exit this region). It follows that in the coordination space, the obstacle region is a rectangular region of length $l_j+L_i$ along axis $i$ and $l_i+L_j$ along axis $j$ (see the top right drawing of Figure~\ref{fig-example-collision}).
\label{ex:collision1}
\end{example}

\begin{example}[Two robots along the same straight path]
Consider two robots of length $L$ traveling along the same straight paths as depicted in the bottom part of Figure~\ref{fig-example-collision} and assume that the same parametrization of geometric paths is used for both robots, i.e., $\gamma_i(x_i)=\gamma_j(x_i)$ if and only if $x_i=x_j$. There are two options: either robot $i$ follows robot $j$ and collision avoidance requires $x_j\geq x_i+L$, or robot $j$ follows robot $i$ and collision avoidance requires $x_i \geq x_j+L$. Hence, the collision avoidance requirement including both cases is: $\vert x_i-x_j\vert \geq L$, and the obstacle region is the band $\left\{ x\in(0,1)^\robots: \vert x_i-x_j\vert < L\right\}$. 
\label{ex:collision2}
\end{example}

\subsection{Priorities: definition and properties}

This subsection shows that the intuitive notion of "assigning priorities" is equivalent to a completion of the obstacle region. It is indeed equivalent to consider as forbidden configurations both collision configurations and configurations that do not respect the assigned priorities, resulting in a completed obstacle region. 

Let $\chiobs_{i\succ j} $ and $\chifree_{i \succ j}$ denote the subsets of $\chi$ defined below:
\begin{eqnarray}
\chiobs_{i\succ j} &:= &(\chiobs_{ij} - \RR_+ \e_i + \RR_+ \e_j)\cap\chi
\label{eq-fixed-priority-collision-cylinder}\\
\chifree_{i \succ j}&:=&\chi \setminus \chiobs_{i\succ j}
\end{eqnarray}
\begin{figure}[!htbp]
\begin{center}
\includegraphics[width=0.7\linewidth]{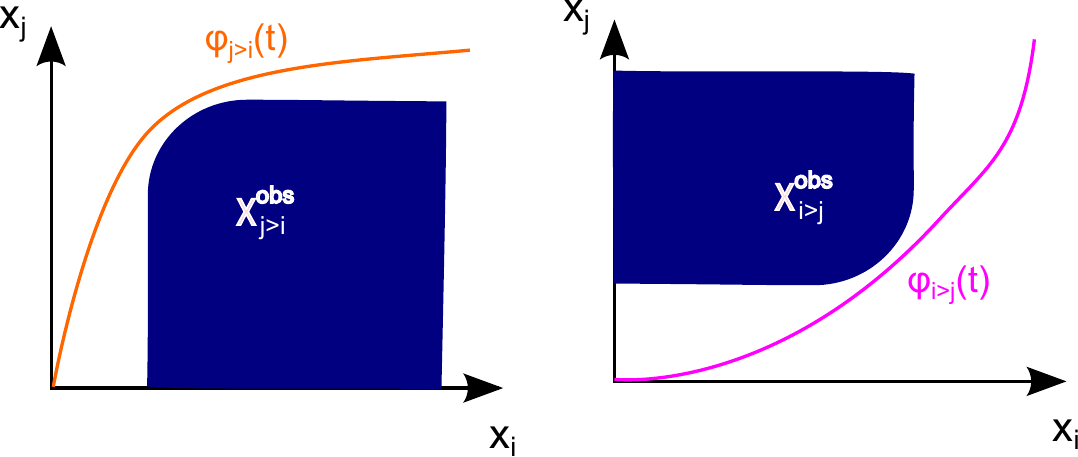}\hfill
\end{center}
\caption{Projection of the completed collision cylinders $\chiobs_{i\succ j}$ and $\chiobs_{j\succ i}$. In this example, the path $\path_{i\succ j}$ (resp. $\path_{j\succ j}$) is collision-free with $\chiobs_{i\succ j}$ (resp. $\chiobs_{j\succ i}$), so the induced priority relation satisfies $i \succ j$ (resp. $j \succ i$).}
\label{fig-fixed-priority-obstacle-region}
\end{figure}

Figure~\ref{fig-fixed-priority-obstacle-region} displays the sets $\chiobs_{i\succ j}$ and $\chiobs_{j \succ i}$. The rationale behind the definition of these sets is that as a feasible path is non-decreasing, it necessarily lies below or above each collision cylinder as depicted in Figure~\ref{fig-fixed-priority-obstacle-region}. This reflects the intuitive notion of priority at intersections. Deciding on which side to pass with respect to each collision cylinder is equivalent to deciding the relative order of robots to go through the intersection. The two geometric invariance properties that follow will be useful to prove results in the sequel.

\begin{property}[Geometric invariances of $\chiobs_{i\succ j}$ and $\chifree_{i\succ j}$] For all $i,j\in\robots$, the following identities hold:
\begin{eqnarray}
\left(\chiobs_{i\succ j}-\RR_+\e_i+\RR_+\e_j\right) \cap \chi &=& \chiobs_{i \succ j}\label{eq:geometric-invariance-obstacle}\\
\left(\chifree_{i\succ j}+\RR_+\e_i-\RR_+\e_j\right)\cap\chi &=& \chifree_{i \succ j}\label{eq:geometric-invariance-collision-free}
\end{eqnarray}
\label{property:geometric-invariance}
\end{property}
\begin{property}[Invariance through $\min$ and $\max$ operators]
\label{property:min-max}
Given $x,y\in\chi$, for all $i,j\in\robots$, the following implications hold:
\begin{eqnarray}
x,y\in\chifree_{i\succ j} &\Rightarrow&  \max\{x,y\} \in \chifree_{i\succ j} \label{eq:property-max}\\
x,y\in\chifree_{i\succ j} &\Rightarrow&  \min\{x,y\} \in \chifree_{i\succ j}\label{eq:property-min}
\end{eqnarray}
\end{property}

In the sequel, we are going to show that the definition of the sets $\chiobs_{i\succ j}$ enables to define rigorously the so-called priority relation induced by a feasible path. The definition of the completed obstacle region enables to easily define a priority relation for feasible paths. The fact that a feasible path necessarily and exclusively lies on one side or on the other side of each collision cylinder $\chiobs_{ij}$ is indeed equivalent to intersect, necessarily and exclusively, one of the completed cylinders $\chiobs_{i\succ j}$, or $\chiobs_{j\succ i}$.

\begin{definition}[Priority relation]
The priority relation $\succ$ is a binary relation on the set of robots $\robots$. For all $i,j\in\robots$, $i\succ j$ if $\im{\path} \cap \chiobs_{j\succ i} \neq \emptyset$.
\end{definition}
We say $\succ$ is the priority relation induced by $\path$. The theorem below asserts that the relation $\succ$ satisfies basic properties that one can expect from a "priority relation". More precisely, $\succ$ does not define a priority relation between two robots that cannot collide ($\chiobs_{ij}=\emptyset$) and if two robots can potentially collide, a priority relation exists and we have $i\succ j$ or $j\succ i$ exclusively, i.e., if robot $i$ has priority over robot $j$ then robot $j$ does not have priority over robot $i$.

\begin{theorem}[Priority relation properties]
Let $\path\in\phifree$ denote a feasible path and $\succ$ the priority relation induced by $\path$. For all $i,j\in\robots$ such that $\chiobs_{ij}\neq\emptyset$, we have necessarily and exclusively $i \succ j$ or $j \succ i$. For all $i,j\in\robots$ such that $\chiobs_{ij}=\emptyset$, we have $i \not\succ j$.
\label{thm:priority-relation}
\end{theorem}
\begin{proof}
The proof is based on the following lemma, which is well-known and referred to as South-West completion~\cite{ODonnell1989, Gregoire2014-thesis}.
\begin{lemma}[South-West and North-East completion~\cite{ODonnell1989}]
For all feasible paths $\path\in\phifree$,
\begin{equation}
\forall i,j\in\robots, \im{\path}\cap\left(\chiobs_{i\succ j}\cap\chiobs_{j\succ i}\right) = \emptyset
\end{equation}
\label{lemma:south-west-north-east-completion}
\end{lemma}
Now, take a feasible path $\path\in\phifree$ and let $\succ$ denote the priority relation induced by $\path$. Take $i,j\in\robots$ such that $\chiobs_{ij}=\emptyset$. Then, we have $\chiobs_{j\succ i}=\emptyset$, so that $\im{\path}\cap\chiobs_{j\succ i}=\emptyset$, that is $i \not\succ j$. Take $i,j\in\robots$ such that $\chiobs_{ij}\neq\emptyset$ and take $y\in\chiobs_{ij}$. Remember that $\path$ is non-decreasing with $\path(0) = (0\cdots 0) \in \chifree$ and $\path(1)=(1\cdots 1)\in \chifree$. As $y\in(0,1)^\robots$, there are two options:
\begin{enumerate}[(a)]
\item either $y\in(\im{\path}-\RR_+\e_i+\RR_+\e_j)\cap \chi$: it implies that $\im{\path} \cap \chiobs_{j \succ i} \neq \emptyset$;
\item or $y\in(\im{\path}-\RR_+\e_j+\RR_+\e_i)\cap\chi$: it implies that $\im{\path} \cap \chiobs_{i \succ j} \neq \emptyset$.
\end{enumerate}
Hence, a feasible path necessarily intersects $\chiobs_{i\succ j}$ or $\chiobs_{j \succ i}$, so we have necessarily $i \succ j$ or $j \succ i$.

Now, we will prove that it is exclusive by contradiction. Take a feasible path $\path$ and assume that for some $t^1\in[0,1]$, $\path(t^1)\in\chiobs_{i \succ j}$ and for some $t^2\in[0,1]$, $\path(t^2)\in\chiobs_{j \succ i}$. Assume arbitrarily that $t^1\leq t^2$ (otherwise, exchange the roles of $i$ and $j$), which implies that $\path(t^1)\leq\path(t^2)$. 

Using monotonicity of $\path$, we easily obtain that for all $t\in[t^1,t^2]$, $\path(t)\in\chiobs_{i\succ j}\cup\chiobs_{j\succ i}$. If $\path(t)\in\chiobs_{i\succ j}\cap\chiobs_{j \succ i}$ for some $t\in[t^1,t^2]$, $\path$ would not be feasible by Lemma~\ref{lemma:south-west-north-east-completion}. Hence, we have:
\begin{eqnarray}
\path(t^1)&\in&\chiobs_{i\succ j}\setminus\chiobs_{j \succ i} \label{eq:phi-t1}\\
\path(t^2)&\in&\chiobs_{j\succ i}\setminus\chiobs_{i \succ j}\label{eq:phi-t2}
\end{eqnarray}
and for all $t\in[t^1,t^2]$,
\begin{equation}
\path(t)\in  \left(\chiobs_{i\succ j}\setminus\chiobs_{j \succ i}\right) \cup \left(\chiobs_{j\succ i}\setminus\chiobs_{i \succ j}\right)  
\label{eq:phi-in-union-two-cylinders}
\end{equation}
As $\left(\chiobs_{i\succ j}\setminus\chiobs_{j \succ i}\right)\cap\left(\chiobs_{j\succ i}\setminus\chiobs_{i \succ j}\right)=\emptyset$, by continuity of $\path$, there exists some $t^0\in[t^1,t^2]$ such that:
\begin{equation}
\path(t^0)\in \partial \left(\chiobs_{i\succ j}\setminus\chiobs_{j \succ i}\right) \cap \partial \left(\chiobs_{j\succ i}\setminus\chiobs_{i \succ j}\right) = \emptyset
\label{eq:phi-at-frontier-two-cylinders}
\end{equation}
This contradiction concludes the proof.
\end{proof}

As any binary relation, the priority relation admits a graph representation. 
\begin{definition}[Priority graph]
The priority graph induced by a feasible path $\path$ is the oriented graph $G$ whose vertices are $V(G):=\robots$ and such that there is an edge from $i$ to $j$ if $i \succ j$ where $\succ$ denotes the priority relation induced by $\path$. We write $(i,j)\in E(G)$ where $E(G)$ denotes the edge set of the priority graph.
\end{definition}
We let $\Gamma$ denote the application that returns the priority graph $\Gamma(\path)$ induced by a feasible path $\path\in\phifree$. $\Gamma(\path)$ is the graph of the priority relation $\succ$ induced by $\path$. Theorem~\ref{thm:priority-relation} can be rewritten as follows: $\Gamma(\path)\in\graphs$ for all feasible paths $\path\in\phifree$
where $\graphs$ is the set of oriented graphs $G$ with vertices $V(G):=\robots$, whose edge set $E(G)$ satisfies:
\begin{equation}
\forall i,j\in\robots,\quad (i,j)\in E(G) \Leftrightarrow
\left\{\begin{matrix}
 \chiobs_{ij} &\neq& \emptyset\\
(j,i)&\notin& E(G)
\end{matrix}\right.
\end{equation}

We say a graph $G$ is a priority graph if $G\in\graphs$. It is natural as a graph $G\in\graphs$ defines a binary relation between robots whose paths intersect, i.e., it defines a priority between all and only robots that need to coordinate.

\section{Priority encoded homotopy classes}
\label{sec:priority-encoded-homotopy-classes}

\subsection{Priority encoding of navigation homotopy classes}

If previous work already noticed the existence of homotopy classes in multi robot coordination~\cite{Ghrist2005,Ghrist2006}, to our knowledge, no meaningful representative is proposed to encode homotopy classes. In the following, we present the main result of this section: priorities uniquely encode homotopy classes of feasible paths in the coordination space. The existence of a finite number of homotopy classes thus merely appears as the consequence of the finite number of possible priority graphs.

\begin{figure}[!htbp]
\begin{center}
\includegraphics[width=0.8\linewidth]{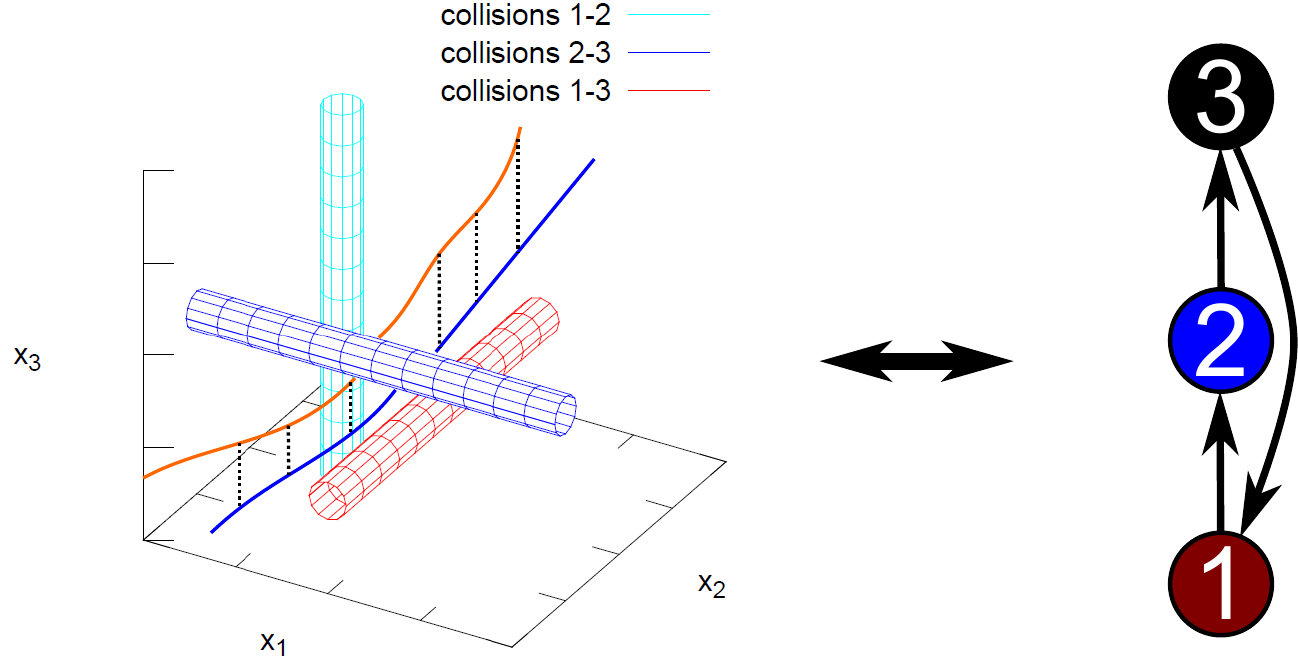}
\end{center}
\caption{A homotopy class of feasible paths in a three-dimensional coordination space and its corresponding unique representative as a priority graph.}
\label{fig:homotopy-priority-graph-unique-representative}
\end{figure}

We let $\Gamma(\phifree):=\{\Gamma(\path):\path\in\phifree\}$ denote the set of values taken by the priority graph over all feasible paths. $\Gamma(\phifree)$, that we will refer to as \emph{feasible priority graphs}, is a subset of $\graphs$ containing graphs $G$ such that there exists a feasible path $\path\in\phifree$ satisfying $\Gamma(\path)=G$. The following theorem (illustrated in Figure~\ref{fig:homotopy-priority-graph-unique-representative}) shows that priorities and homotopy classes are strongly linked: more precisely, there is a bijective relationship between homotopy classes and feasible priority graphs. Note that \emph{this result is compatible with the case of two robots $i,j$ which follow each other}, as such a couple of robots is treated just like robots following different paths (see Example~\ref{ex:collision2}). Robot $i$ following robot $j$ is equivalent to $j\succ i$; robot $j$ following robot $i$ is equivalent to $i \succ j$.

\begin{theorem}[Invariance of the priority graph]
The priority graph is an invariant of the homotopy classes of feasible paths that it is distinct for each class: homotopy classes are in bijection with feasible priority graphs.
\label{thm:invariance-priority-graph}
\end{theorem}
\begin{proof}[Proof of invariance]
First we will prove that the priority graph is an invariant of the homotopy classes of feasible paths. Consider a feasible path $\path\in\phifree$. For all $i,j\in\robots$, $(i,j)\in E(\Gamma(\path))$ if $\path$ intersects $\chiobs_{j \succ i}$ and the set $\chiobs_{j \succ i}$ is open. If a feasible path $\path$ intersects an open set, any feasible path $\psi\in\phifree$ close enough to $\path$ (in the topology of pointwise convergence) also intersects this open set. Hence, we have:
\begin{equation}
\forall i,j\in\robots,~ (i,j)\in E(\Gamma(\path)) \Leftrightarrow (i,j)\in E(\Gamma(\psi))
\end{equation}
provided $\psi$ is close enough to $\path$. Therefore, $\Gamma$ is continuous and since it takes discrete values, it is thus constant in homotopy classes of feasible paths. (We identify $\Gamma$ with the set of applications $g_{ij}:\paths(\chifree)\to\{-1,0,1\}$ satisfying $g_{ij}(\path)=1$ if $i\succ j$, $-1$ if $j\succ i$, and $0$ otherwise.) In conclusion, the priority graph is an invariant of the homotopy classes of feasible paths.
\end{proof}
\begin{proof}[Proof of uniqueness]
To prove uniqueness, consider two feasible paths $\path^1$ and $\path^2$ with the same induced priority graph $G$: $\path^1,\path^2\in\phifreeG$. We have to prove that $\path^1$ and $\path^2$ are homotopic. Consider the following continuous transformation: 
\begin{equation}
H: \alpha\in[0,1] \mapsto \min\left\{ \path^1(\bullet+\alpha), \max\left\{ \path^1, \path^2 \right\} \right\}
\end{equation}
where by convention $\path^1(t+\alpha)= \path^1(1)$ if $t+\alpha\geq 1$. Figure~\ref{fig:homotopy-example} illustrates the proposed transformation in the particular case where the two paths have the same endpoints.

\begin{figure}
\begin{center}
\includegraphics[width=1.0\linewidth]{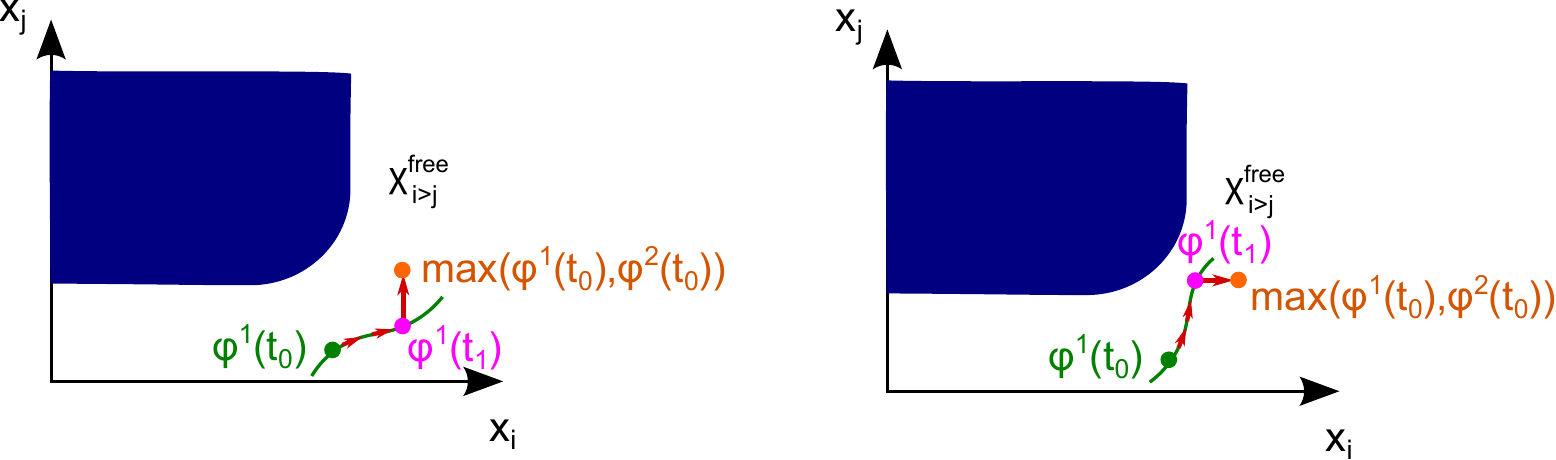}
\end{center}
\caption{Illustration of the transformation of $\path^1$ into $\max(\path^1,\path^2)$. At any point of time $t_0$, $\max(\path^1(t_0),\path^2(t_0))$ necessarily lies on the north-east with respect to $\path^1(t_0)$. As a consequence, the two above cases may appear, and in each case, $\path^1(t_0)$ can be continuously transformed into $\max(\path^1(t_0),\path^2(t_0))$ without collision by following the red arrows.}
\label{fig:homotopy-example}
\end{figure}

$H$ is continuous,
\begin{eqnarray}
H(0) &=& \min\left\{ \path^1, \max\left\{ \path^1, \path^2 \right\} \right\} = \path^1\\
H(1) &=& \min\left\{ \path^1(1), \max\left\{ \path^1, \path^2 \right\} \right\} = \max\left\{ \path^1, \path^2 \right\} 
\end{eqnarray}
Hence, $H$  continuously transforms $\path^1$ into $\max\{ \path^1, \path^2 \}$. Now, we prove that for all $\alpha\in[0,1]$, $H(\alpha)$ is a feasible path. We need to prove that for all $\alpha\in[0,1]$, (\ref{item:continuous}) $H(\alpha)$ is continuous, (\ref{item:start-configuration}) satisfies $H(\alpha)(0)=(0\cdots 0)$ and (\ref{item:end-configuration}) $H(\alpha)(1)=(1\cdots 1)$, (\ref{item:non-decreasing}) is non-decreasing, and (\ref{item:collision-free}) is collision-free. 
\begin{enumerate}[(a)]
\item $H(\alpha)$ is continuous as the result of the application of continuous operators $\min$, $\max$ and delay on continuous paths. \label{item:continuous}
\item $\path^1$ and $\path^2$ being feasible, we have $\path^1(0)=(0\cdots 0)$ and $\path^2(0)=(0\cdots 0)$. Hence, we also have $\max(\path^1(0),\path^2(0))=(0\cdots 0)$ and $H(\alpha)(0)=(0\cdots 0)$. \label{item:start-configuration}
\item $\path^1$ and $\path^2$ being feasible, we have $\path^1(1)=(1\cdots 1)$ and $\path^2(1)=(1\cdots 1)$. Hence, we also have $\max(\path^1(1),\path^2(1))=(1\cdots 1)$ and $H(\alpha)(1)=(1\cdots 1)$. \label{item:end-configuration}
\item $H(\alpha)$ is non-decreasing as the result of the application of non-decreasing operators $\min$ and $\max$ on non-decreasing paths.
\label{item:non-decreasing}
\item Take $(i,j)\in E(G)$ and $\alpha,t\in[0,1]$. We have $\path^1(t+\alpha)\in\chifree_{i\succ j}$ as $\path^1\in\phifreeG$ and we have also $\max\{\path^1(t),\path^2(t)\}\in\chifree_{i\succ j}$ as $\path^1,\path^2\in\phifreeG$ and using Property~\ref{property:min-max}. By Property~\ref{property:min-max}, applying the $\min$ operator on two configurations in $\chifree_{i\succ j}$ returns a configuration in  $\chifree_{i \succ j}$. In conclusion, we have $H(\alpha)(t)\in\chifree_G\subset\chifree$
\label{item:collision-free}
\end{enumerate}
As a result, $\path^1$ is homotopic to $\max\{ \path^1, \path^2 \} $. As $\path^1$ and $\path^2$ have symmetric roles, $\path^2$ is homotopic to $\max\{ \path^2, \path^1 \} = \max\{ \path^1, \path^2 \}$ as well. Homotopy defining an equivalence relation, $\path^1$ and $\path^2$ are homotopic.
\end{proof}

\subsection{Guarantees provided by acyclic priorities}

We have proved that all feasible paths sharing the same priorities are continuously deformable into each other forming a homotopy class. A natural question is: does any priority graph $G\in\graphs$ encode a (non-empty) homotopy class of feasible paths ? This paper will only focus on the case of acyclic priorities. Provided the priority graph is acyclic, it is guaranteed that there exists a non-empty homotopy class encoded by the given priorities as stated and proved in the theorem below.

\begin{theorem}[Sufficient condition for priorities feasibility]
All acyclic priority graphs are feasible, i.e., given an acyclic priority graph $G$, there exists a feasible path $\path\in\phifree$ satisfying $\Gamma(\path)=G$.
\label{thm:acyclic-priorities}
\end{theorem}
\begin{proof}
Take an acyclic priority graph $G\in\graphs$. To prove that $G$ is feasible, we are going to exhibit a particular feasible path whose induced priority graph is $G$. As $G$ is acylic, it admits a topological ordering of its nodes $\robots$. Consider a relabeling of robots along this topological ordering, i.e., robot $1$ is the maximal element of this topological ordering, ... robot $i$ is the $i$th element of the topological ordering, ... and robot $n$ is the minimal element of the topological ordering. Consider the path $\path$ constructed as follows. $\path(0):=0$ and for all $i\in\{1\cdots n\}$, within time interval $[(i-1)/n, i/n]$, robot $i$ moves forward from $0$ to $1$ (for example $\path_i$ is linear in that time interval and takes values $[0,1]$) while other robots $j\neq i$ do not move ($\path_j$ constant in that time interval). This path is feasible and takes values in $\chifree_G$.
\end{proof}

\section{A safe way to navigate within homotopy classes under kinodynamic constraints}
\label{sec:control}

The two previous sections analyzed the set of solutions to the coordination problem and described the homotopy structure using priorities. In contrast, this section is constructive. We consider an \emph{acyclic priority graph} $G$ as given, encoding a (non-empty) assigned homotopy class by Theorem~\ref{thm:acyclic-priorities}, and we design a control scheme guaranteeing that the resulting trajectory of robots in the coordination space remains within the assigned homotopy class (first introduced in our preliminary conference paper~\cite{Gregoire2013-dynamic-constraints}).

\subsection{Navigation with kinetic constraints}

In the absence of inertia, navigation within an assigned homotopy class can be achieved easily by letting robots travel at maximum speed, stopping just in time to respect priorities, as described in~\cite{Ghrist2006, Gregoire2014-thesis}. Moreover, the resulting path in the coordination space -- referred to as left-greedy  -- is provably time-optimal as proved in~\cite{Ghrist2006} using tools of CAT(0) geometry~\cite{Bridson1999}. Our priority-based approach provides a new view of left-greedy paths in the coordination space. They are indeed optimal paths within an assigned homotopy class and can now thus be seen as optimal paths under assigned priorities. Difficulty in navigating within an assigned homotopy class arises when considering inertia, so that robots cannot stop instantly to avoid collisions. 

\subsection{Navigation with kinodynamic constraints}

We consider a simple control model, assuming the acceleration of robots can be directly controlled. The proposed control scheme is inspired by References~\cite{DelVecchio2009,Colombo2012}. The key difference is that by introducing homotopic constraints, robots retain a large freedom of action to handle unpredicted events as highlighted by the two propositions at the end of the section and illustrated through simulations.

\subsubsection{The multiple robot system as a monotone controlled system}

Each robot $i$ is modelled as a second-order controlled system with state $s_i=(x_i,v_i)\in S_i:=\RR \times [0,\vmax_i]$, whose evolution is described by the differential equation:
\begin{eqnarray}
\dot{x_i}(t) &= &  1_{x_i(t)< 1}~v_i(t)
\label{eq-diff-state1-deterministic}
\\
\dot{v_i}(t) & =& \ubf_i(t) ~\delta(\ubf_i(t),v_i(t))
\label{eq-diff-state2-deterministic}
\end{eqnarray}
where $\ubf_i:\RR_+ \to U_i$ is the control of robot $i$ and $\vmax_i$ denotes the non-negative speed limit for robot $i$. We let $U_i:=[\uL_i,\uH_i]$ be the set of feasible control values. $\uL_i<0$ represents the maximum brake control value and $\uH_i>0$ represents the maximum throttle control value. $\delta$ is a binary function merely ensuring that $v_i\in [0,\vmax_i]$ at all times, that is, $\delta(\ubf_i(t),v_i(t))=1$ except for $v_i(t)=0$ and $\ubf_i(t)<0$, and for $v_i(t)=\vmax_i$ and $\ubf_i(t)>0$, where it vanishes. The binary multiplicative term $1_{x_i(t) < 1}$ ensures $x_i(t)\leq 1$ by vanishing if and only if $x_i(t) \geq 1$.

The control is assumed to be updated in discrete time every $\dt>0$: 
\begin{equation}
\forall k\in\NN, \forall t\in[k\dt, (k+1)\dt), \ubf_i(t) = \ubf_i(k\dt)
\end{equation}
The time interval $[k\dt, (k+1)\dt)$ will be referred to as (time) slot $k$. For the sake of simplicity we let $\dt := 1$ in the sequel. We let $\controls_i$ denote the set of controls $\ubf_i:\RR_+ \to U_i$ piecewise constant on intervals $[k,k+1)$, $k\in\NN$. We let $t \mapsto \Phi_i(t,s_i,\ubf_i)$ denote the flow of the system starting at initial condition $s_i\in S_i$  with control $\ubf_i \in \controls_i$. 

We also define the vectorial state $s:=(s_i)_{i\in\robots}\in S$, the vectorial control $\ubf:=(\ubf_i)_{i\in\robots}\in \controls:=\prod_{i\in\robots}\controls_i$, and the vectorial flow: $\Phi(t,s,\ubf):=(\Phi_i(t,s_i,\ubf_i))_{i\in\robots}$. We let $\uL:=(\uL_i)_{i\in\robots}$, $\uH:=(\uH_i)_{i\in\robots}$ and we define the constant controls $\uLbf(t):=\uL$ and $\uHbf(t):=\uH$. We introduce partial orders as follows:
\begin{eqnarray}
\forall \ubf_i^1,\ubf_i^2\in \controls_i, \ubf_i^1 \preceq \ubf_i^2 &\text{if}& \forall t\geq 0, \ubf_i^1(t) \leq \ubf_i^2(t)\\
\forall s_i^1=(x_i^1,v_i^1),s_i^2=(x_i^2,v_i^2)\in S_i, s_i^1 \preceq s_i^2 &\text{if}& x_i^1 \leq x_i^2 \text{ and } v_i^1 \leq v_i^2\label{orderr:eq}\\
\forall \Phi^1,\Phi^2:\RR_+\to S, \Phi^1 \preceq \Phi^2 &\text{if}& \forall t\geq 0,\Phi^1(t) \preceq \Phi^2(t)
\end{eqnarray}
The controlled system~\eqref{eq-diff-state1-deterministic}-\eqref{eq-diff-state2-deterministic}  is a monotone control system \cite{Angeli2003} with regards to the relative orders defined above. More precisely, the following key property holds:

\begin{property}[Order preservation]
The flow $t \mapsto \Phi_i(t,s_i,\ubf_i)$ is order-preserving with regards to $s_i$ and $\ubf_i$.
\end{property}

\subsubsection{The proposed control law}

\label{law:sec}
We define projection operators as follows: $\pi_x(s):=x$ and  $\pi_{x,i}(s):=\pi_{x,i}(s_i):=x_i$. $G$ denotes a given priority graph. 
Define the set of brake safe states as follows:
\begin{equation}
B_{G}:=\{ s\in S: \pi_x\left(\Phi\left(\RR_+,s,\uLbf\right)\right) \subset \chifree_G\} \subset S
\end{equation} 
According to the above definition, a state $s\in S$ is brake safe if, starting at initial condition $s$ under maximum brake control, the system remains in $\chifree_G$. In particular, a state $(x,0)$ with $x\in\chifree_G$ is brake safe, so $B_G$ is not empty provided $\chifree_G$ is not empty. We now build a control law $g^G:S\to U$ such that starting from an initial brake safe state in $B_G$, the flow of the system controlled by the control law $g^G$ is ensured to remain in $B_G$ (thus being collision-free and respecting priorities $G$). In other words, $B_G$ shall be positively invariant for the system under control law $g^G$. 

The rationale for our control law is as follows. Consider a robot $i$ and a robot $j$ that has priority over $i$. Given an initial configuration of the two robots, the worst-case scenario is when $j$ brakes whereas $i$ accelerates in the next time slot. If the trajectory of the system in the next time slot under that worst-case scenario is collision-free and if the reached state is brake safe, robot $i$ may accelerate in any case. Otherwise, it is required to brake. This is formalized below.

Let $\ubf_i^\imp\in \controls_i$ denote the impulse control for robot $i$ and $\tilde{\ubf}^i$ denote the worst-case vectorial control with regards to $i$, defined as follows:
\begin{eqnarray}
\ubf_i^\imp(k) &:=& 
\begin{cases}
\uH_i & \text{if } k=0\\
\uL_i & \text{if } k\geq 1
\end{cases}
\label{eq:impulse-control}\\
\tilde{\ubf}^i_j&:=&\begin{cases}
 \ubf_i^\imp & \text{if } j=i\\
\uLbf_j & \text{if } j\neq i
\end{cases}
\end{eqnarray}

The control law can then be formulated synthetically:
\begin{equation}
g_i^G(s):=\begin{cases}
\uL_i & \text{if } \exists (j,i)\in E(G) \text{ s.t. } \pi_x(\Phi(\RR_+,s,\tilde{\ubf}^i)) \cap \chiobs_{j\succ i} \neq \emptyset\\
\uH_i & \text{ else.}
\end{cases}
\label{eq-control-map}
\end{equation}
This simply means that robot $i$ applies maximum throttle command unless the worst-case flow $t\mapsto\Phi(t,s,\tilde{\ubf}^i)$ intersects $\chiobs_G$ at some point of time $t\geq 0$, in which case it applies maximum brake command.

\subsubsection{Safety, robustness and liveness properties of the proposed control law}

The theorem below asserts that the control law $g_i^G$ returns the maximum control value that robot $i$ can safely apply to remain brake safe. Provided the system starts in a brake safe state, the sequence of future states at the beginning of each time slot is a sequence of brake safe states (see Equation~\eqref{eq:invariance-B-G}) as long as Inequality~\eqref{eq-inequality-control-map} is satisfied. Moreover, the flow of the system remains in $\chifree_G$ in continuous time (see Equation~\eqref{eq:no-collision-during-time-slots}), i.e., no collision occurs and priorities are preserved.

\begin{theorem}[A broad class of priority-preserving controls]
Given an initial condition $s\in B_G$, and a control $\ubf \in \controls$ that satisfies: 
\begin{equation}
\forall k\in\NN, \ubf(k) \leq g^G(\Phi(k,s,\ubf))
\label{eq-inequality-control-map}
\end{equation}
The set of brake safe states $B_G$ is positively invariant (in discrete time), i.e.:
\begin{equation}
\forall k\in\NN, \Phi(k,s,\ubf) \in B_G
\label{eq:invariance-B-G}
\end{equation}
Moreover, the configuration of the system remains in $\chifree_G$ through time, i.e.:
\begin{equation}
\forall t\geq 0, \pi_x(\Phi(t,s,\ubf))\in\chifree_G
\label{eq:no-collision-during-time-slots}
\end{equation}
\label{thm-robustness-brake-application}
\end{theorem}

\begin{proof}
For two first statements, by induction, it is sufficient to prove that given an initial condition $s\in B_G$, the flow is collision-free for $t\in[0,1]$ and the reached state $\Phi(1,s,\ubf)$ is brake safe. We begin with the proof that the flow of Theorem~\ref{thm-robustness-brake-application} does not intersect $\chiobs_G$ for $t\in[0,1]$. Take arbitrary $t\in[0,1]$: we have to prove that for all $(j,i)\in E(G)$, $\pi_x(\Phi(t,s,\ubf))\in\chifree_{j \succ i}$. By construction of $g^G$, for each robot $i$, there are two cases:
\begin{itemize}
\item $g_i^G(s)=\uL_i$: in this case,
\begin{equation}
\Phi_i(t,s,\ubf)=\Phi_i(t,s,\uLbf)
\label{eq:pf-discrete-case1-eq1}
\end{equation}
and by order-preservation, for all robots $j$ such that $(j,i)\in E(G)$ we have: 
\begin{equation}
\Phi_j(t,s,\ubf) \geq \Phi_j(t,s,\uLbf)
\label{eq:pf-discrete-case1-eq2}
\end{equation}
Since $s$ is brake safe, $\pi_x (\Phi(t,s,\uLbf))\in\chifree_{j \succ i}$. Hence, by Property~\ref{property:geometric-invariance}, Equations~\eqref{eq:pf-discrete-case1-eq1} and~\eqref{eq:pf-discrete-case1-eq2} ensure that $\pi_x(\Phi(t,s,\ubf))\in\chifree_{j \succ i}$ as well.
\item $g_i^G(s)=\uH_i$: by construction of the control law, $\pi_x(\Phi(t,s,\tilde{\ubf}^i)) \in \chifree_{G}$. By order-preservation, using $\tilde{\ubf}^i_i(0)=\uH_i$, we obtain:
\begin{equation}
\Phi_i(t,s,\tilde{\ubf}^i)=\Phi_i(t,s,\uHbf) \geq \Phi_i(t,s,\ubf)
\label{eq:pf-discrete-case2-eq1}
\end{equation}
For all robots $j$ such that $(j,i)\in E(G)$, using $\tilde{\ubf}^i_j(0)=\uL_j$, we have:
\begin{equation}
\Phi_j(t,s,\tilde{\ubf}^i)=\Phi_j(t,s,\uLbf) \leq   \Phi_j(t,s,\ubf)
\label{eq:pf-discrete-case2-eq2}
\end{equation}
Since $\pi_x(\Phi(t,s,\tilde{\ubf}^i)) \in \chifree_{G}$, $\pi_x(\Phi(t,s,\tilde{\ubf}^i))\in\chifree_{j \succ i}$, and by Property~\ref{property:geometric-invariance}, Equations~\eqref{eq:pf-discrete-case2-eq1} and~\eqref{eq:pf-discrete-case2-eq2} ensure that $\pi_x(\Phi(t,s,\ubf))\in\chifree_{j \succ i}$ as well.
\end{itemize}

As a final step, let us  prove that the reached state $s^1:=\Phi(1,s,\ubf)$ is brake safe. Take arbitrary $t \geq 0$: we have to prove that for all $(j,i)\in E(G)$, $\pi_x(\Phi(t,s^1,\uLbf))\in\chifree_{j \succ i}$. As previously, there are two cases: 
\begin{itemize}
\item $g_i^G(s)=\uL_i$: then, $s^1_i=\Phi_i(1,s,\uLbf)$ and we have: 
\begin{equation}
\Phi_i(t,s^1,\uLbf) = \Phi_i(1+t,s,\uLbf)
\label{eq:pf-continuous-case1-eq1}
\end{equation}
Moreover, by order-preservation, for all $j$ such that $(j,i)\in E(G)$: $s_j^1 \geq \Phi_j(1,s,\uLbf)$. As a result, by order-preservation:
\begin{equation}
\Phi_j(t,s^1,\uLbf) \geq \Phi_j(1+t,s,\uLbf)
\label{eq:pf-continuous-case1-eq2}
\end{equation}
Since $s$ is brake safe, $\pi_x (\Phi(1+t,s,\uLbf))\in\chifree_{j \succ i}$. Hence, by Property~\ref{property:geometric-invariance}, Equations~\eqref{eq:pf-continuous-case1-eq1} and~\eqref{eq:pf-continuous-case1-eq2} ensure that $\pi_x(\Phi(t,s^1,\uLbf))\in\chifree_{j \succ i}$ as well.

\item $g_i^G(s)=\uH_i$: then, by construction of the control law, $\pi_x(\Phi(1+t,s,\tilde{\ubf}^i)) \in \chifree_G$. Define $\tilde s^1:=\Phi(1,s,\tilde{\ubf}^i)$. We have $\tilde{\ubf}^i(1+\tau)=\uL$ for $\tau\geq 0$. As a result, $\Phi(1+t,s,\tilde{\ubf}^i)=\Phi(t,\tilde{s}^1,\uLbf)$. Since $\pi_x(\Phi(1+t,s,\tilde{\ubf}^i)) \in \chifree_G$,  $\pi_x(\Phi(t,\tilde s^1,\uLbf))\in \chifree_G$.

By order-preservation, using $\tilde{\ubf}^i_i(0)=\uH_i$, we obtain:
\begin{equation}
\tilde s^1_i = \Phi_i(1,s,\tilde{\ubf}^i) = \Phi_i(1,s,\uHbf)  \geq   \Phi_i(1,s,\ubf) = s^1_i
\label{eq:pf-continuous-case2-eq1-intermediate-equation}
\end{equation}
For all robots $j$ such that $(j,i)\in E(G)$, using $\tilde{\ubf}^i_j(0)=\uL_j$, we have:
\begin{equation}
\tilde s^1_j = \Phi_j(1,s,\tilde{\ubf}^i) = \Phi_j(1,s,\uLbf)   \leq \Phi_j(1,s,\ubf) = s^1_j
\label{eq:pf-continuous-case2-eq2-intermediate-equation}
\end{equation}
Hence, by order-preservation, Equations~\eqref{eq:pf-continuous-case2-eq1-intermediate-equation} and~\eqref{eq:pf-continuous-case2-eq2-intermediate-equation} imply:
\begin{eqnarray}
\Phi_i(t,\tilde{s}^1,\uLbf) & \geq & \Phi_i(t,s^1,\uLbf)\label{eq:pf-continuous-case2-eq1}\\
\Phi_j(t,\tilde{s}^1,\uLbf) & \leq & \Phi_j(t,s^1,\uLbf)\label{eq:pf-continuous-case2-eq2}
\end{eqnarray}
Since $\pi_x(\Phi(t,\tilde s^1,\uLbf))\in \chifree_G$, $\pi_x(\Phi(t,\tilde s^1,\uLbf))\in \chifree_{j \succ i}$, and by Property~\ref{property:geometric-invariance}, Equations~\eqref{eq:pf-continuous-case2-eq1} and~\eqref{eq:pf-continuous-case2-eq2} ensure that $\pi_x(\Phi(t,s^1,\uLbf))\in \chifree_{j \succ i}$ as well.
\end{itemize}
\end{proof}

Finally, the proposed control scheme allows all robots to eventually go through the intersection, this is a liveness result stated in the theorem below.

\begin{theorem}[Liveness]
Given an initial condition $s\in B_G$ and a control signal $\ubf \in \controls$ satisfying $\forall k\in\NN, \ubf(k) = g^G(\Phi(k,s,\ubf))$, all robots eventually go through the intersection, i.e, $\pi_x(\Phi(T,s,\ubf)) = (1\cdots 1)$ for some $T\geq 0$.
\label{thm:liveness}
\end{theorem}
\begin{proof}
Consider the trajectory of the robots under control law $g^G$. $G$ being a directed acyclic graph, there exists an extremal vertex $i_1\in\robots$ such that for all $j\in\robots$, $(j,i_1)\notin E(G)$. As a result, under the control law $g^G$, robot $i_1$ will always accelerate as much as possible and it will exit the intersection, i.e., reach position $1$, in finite time $T_1$.

Now, assume that at time $T_m$, robots $i_1 \cdots i_m$ have exited the intersection and $m<n$ (there remain some robots). $G$ being acyclic, there exists an extremal element for the remaining robots denoted $i_{m+1}\in\robots\setminus \{i_1 \cdots i_m\}$ such that for all $j\in\robots\setminus \{i_1 \cdots i_m\}$, $(j,i_{m+1})\notin E(G)$. Collisions occurring only with non exited robots, for $t\geq T_m$ $j$ will always accelerate and it will exit the intersection in finite time at instant $T_{m+1}\geq T_m$. 

Iterating this process for $m=1 \cdots n-1$ yields a sequence $(T_1 \cdots T_n)$ and all robots have exited the intersection at time $T:=T_n$, and $\pi_x(\Phi(T,s,\ubf)) = (1\cdots 1)$.
\end{proof}

The proposed control scheme allows for reactivity because robot $i$ is not required to apply the control value returned by $g_i^G$. All values below the value returned by $g_i^G$ are acceptable, and still guarantee to remain in the assigned homotopy class. This provides freedom of action that can be used to react to unpredicted events as illustrated by the two following propositions and simulations of the next section.

\section{Illustration of the obtained robustness property}
\label{sec:robustness}

This section illustrates our approach's robustness by considering the two following concrete scenarios. 

\begin{proposition}[Robustness to individual brake application]
Given acyclic priorities $G$, an initial brake safe state $s\in B_G$, a particular robot $i\in\robots$ and a finite subset of slots $K \subset \NN$, consider a control $\ubf \in\controls$ satisfying: 
\begin{eqnarray}
\forall k\in\NN,  \ubf_i(k)&=&\begin{cases}
\uL_i & \text{ if } k\in K \\
g_i^G(\Phi(k,s,\ubf)) & \text{ else.}
\end{cases} \label{eq-example-control}\\
\forall j\in\robots, j\neq i, \ubf_j(k)&=&g_j^G(\Phi(k,s,\ubf))
\end{eqnarray}
The trajectory of robots in the coordination space will take values in $\chifree_G$ and all robots will eventually go through the intersection.
\label{cor:individual-brake}
\end{proposition}
Under the control described above, the system is under the control law $g^G$, except during slots $K$ where the particular robot $i$ brakes while other robots $j$ are still under the control law $g_j^G$. Such a scenario may arise, for instance, in case of a momentary communication/sensing failure for one robot. The proof of the above proposition is direct as the condition of Theorem~\ref{thm-robustness-brake-application} is clearly respected since for $j\neq i$, $\ubf_j(k)=g_j^G(\Phi(k,s,\ubf))\leq g_j^G(\Phi(k,s,\ubf))$, and  $\ubf_i(k)=g_i^G(\Phi(k,s,\ubf))\leq g_i^G(\Phi(k,s,\ubf))$ or $\ubf_i(k)=\uL_i \leq g_i^G(\Phi(k,s,\ubf))$. As the system is under control law $g^G$ after time slots $K$, liveness is also guaranteed by Theorem~\ref{thm:liveness}. It illustrates that the control law is robust with regards to an individual brake application of a particular robot for an arbitrary long time.

\begin{proposition}[Robustness to simultaneous brake application]
Given acyclic priorities $G$, an initial brake safe state $s\in B_G$ and a finite subset of slots $K \subset \NN$, consider a control $\ubf \in\controls$ satisfying: 
\begin{equation}
\forall k\in\NN,  \ubf(k)=\begin{cases}
\uL & \text{ if } k\in K \\
g^G(\Phi(k,s,\ubf)) & \text{ else.}
\end{cases} \label{eq-example-control}
\end{equation}
The trajectory of robots in the coordination space will take values in $\chifree_G$ and all robots will eventually go through the intersection.
\label{cor:simultaneous-brake}
\end{proposition}
Under the control described above, the system is under the control law $g^G$, except during slots $K$ where all robots brake simultaneously. It may arise in case of a global failure requiring an emergency brake to be performed. Again, the proof is direct as the condition of Theorem~\ref{thm-robustness-brake-application} is clearly respected since $\ubf(k)=g^G(\Phi(k,s,\ubf)) \leq g^G(\Phi(k,s,\ubf))$ or $\ubf(k)=\uL \leq g^G(\Phi(k,s,\ubf))$. As the system is under control law $g^G$ after time slots $K$, liveness is also guaranteed by Theorem~\ref{thm:liveness}. It illustrates that the control law is robust with regards to a simultaneous brake application of all robots for an arbitrary long time.

\begin{figure}[h]
\begin{center}
\includegraphics[width=0.45\linewidth]{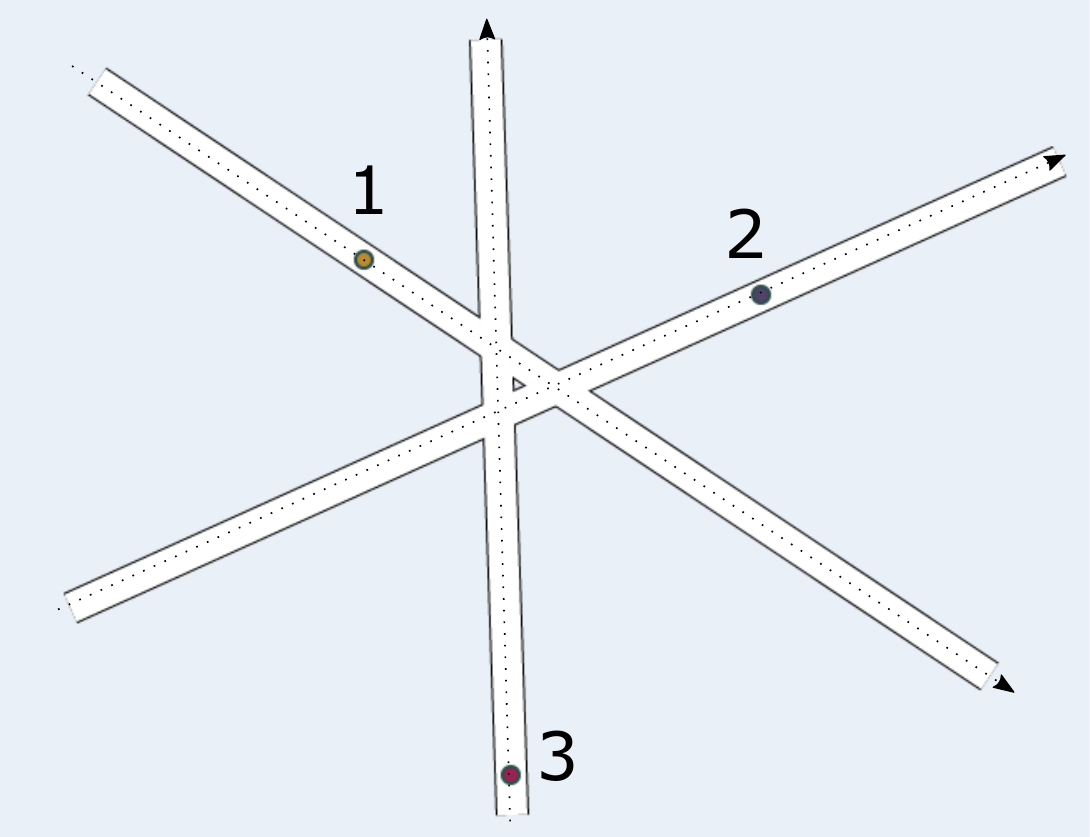}\hfill
\includegraphics[width=0.45\linewidth]{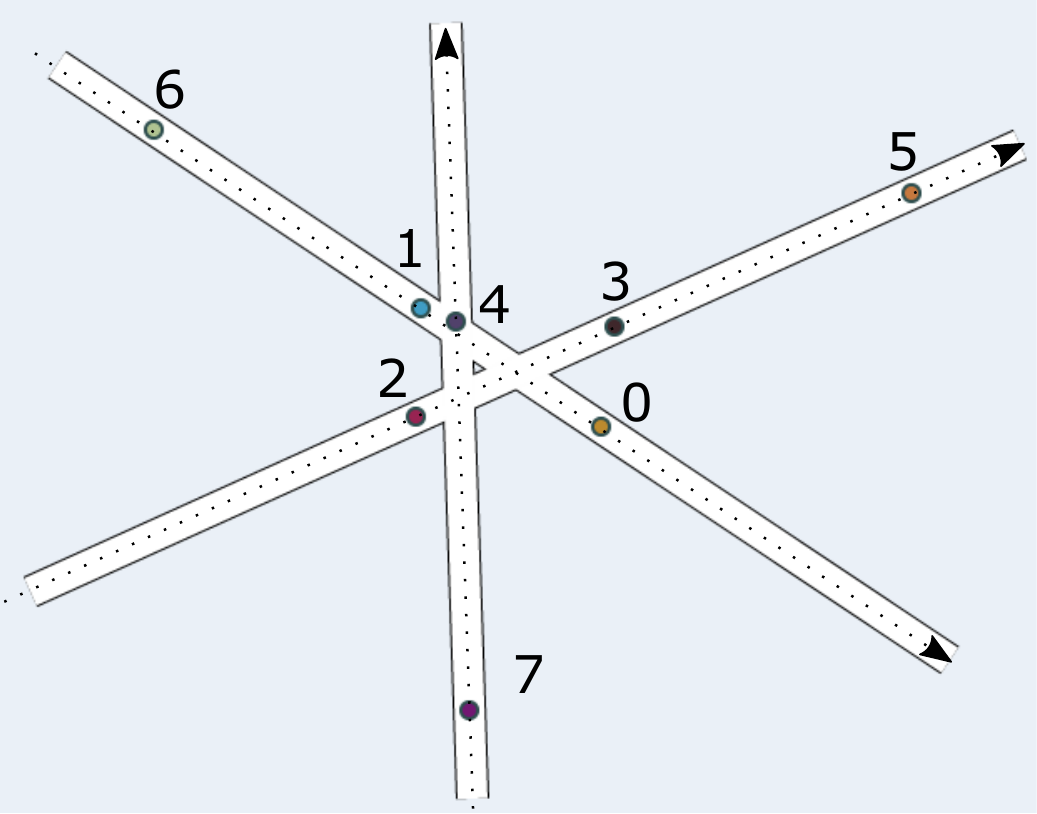}\hfill
\end{center}
\caption{The two initial configurations considered in simulations: three robots in a three-path intersection scenario (left) and height robots in a three-path intersection scenario (right).}
\label{fig:intersection-initial-state}
\end{figure}

\begin{figure}[h]
\begin{center}
\vspace{-2.0cm}
\includegraphics[width=0.9\linewidth]{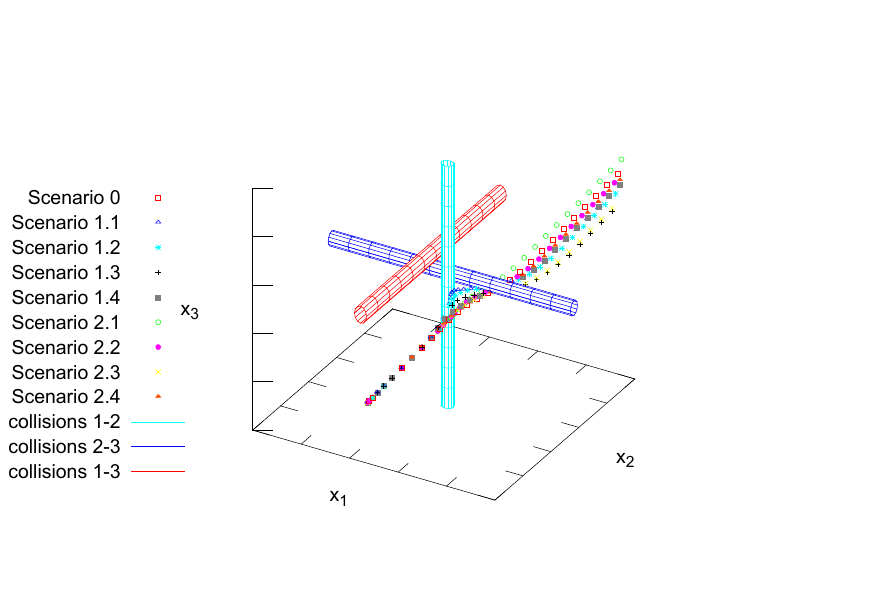}\hspace{-2.5cm}
\vspace{-1.2cm}
\end{center}
\caption{Trajectory described in the coordination space for the 9 scenarios with three robots.}
\label{fig:trajectories}
\end{figure}

Simulations have been conducted to illustrate the robustness of our approach and the role of priority encoded homotopic constraints. For the sake of the simplicity, the implementation (in Java) considers circle-shaped robots with a common diameter $D$ along straight paths. The lateral control is not simulated, robots being assumed to follow their assigned geometric path. At maximum velocity, robots travel distance $D/2$ during one slot. They share the same kinodynamic constraints with $\uL=-\uH$ and 20 slots are necessary to go from stop to full speed (and conversely). We have considered the three-path intersection depicted in Figure~\ref{fig:intersection-initial-state}. In the first simulation set, there is one robot on each path as depicted in Figure~\ref{fig:intersection-initial-state} (left). Robots $1$ and $2$ are at position $0.3$ and robot $3$ is at $0.05$. They are all initially stopped. Priorities are $1\succ 2\succ 3$ and $1\succ 3$. We have conducted 9 simulation runs. In scenario 0, robots are under control law $g^G$; a video capture of this scenario is available \href{https://www.youtube.com/watch?v=hAILADwBIbA}{here}\footnote{\url{https://www.youtube.com/watch?v=hAILADwBIbA}}. In scenarios 1.1, 1.2, 1.3 and 1.4, the control signal satisfies the condition of Proposition~\ref{cor:individual-brake} where robot $0$ is the robot braking unexpectedly, with respectively $K=[25,45]$, $K=[25,40]$, $K=[25,35]$ and $K=[25,30]$; video capture of scenario 1.1 \href{https://www.youtube.com/watch?v=CLM1RfV3vg4}{here}\footnote{\url{https://www.youtube.com/watch?v=CLM1RfV3vg4}}. In scenarios 2.1, 2.2, 2.3 and 2.4, the control signal satisfies the condition of Proposition~\ref{cor:simultaneous-brake} with respectively $K=[25,45]$, $K=[25,40]$, $K=[25,35]$ and $K=[25,30]$; video capture of scenario 2.1 \href{https://www.youtube.com/watch?v=hGy_pPTNszw}{here}\footnote{\url{https://www.youtube.com/watch?v=hGy_pPTNszw}}. The path described in the coordination space is depicted for all scenarios in Figure~\ref{fig:trajectories}. All described paths are clearly homotopic to each other. One can see that the path in the coordination space can much deviate depending on unexpected events, yet remaining in the same homotopy class. 

We have also conducted simulations with more robots to illustrate the approach in a higher dimensional coordination space. The intersection area is the same, there are however 8 robots now. As depicted in Figure~\ref{fig:intersection-initial-state} (right), robots $0$, $1$ and $2$ are at position $0.6$; robots $3$ and $4$ are at position $0.4$; and robots $5$, $6$ and $7$ are at position $0.1$. All robots are initially at maximum velocity and they are labeled by descending priorities, i.e., $i \succ j$ if and only if $i < j$. As the coordination space is 8-dimensional, the trajectory described in the coordination space cannot be represented. However, a video capture is available \href{https://www.youtube.com/watch?v=j_3aYj5Hehk}{here}\footnote{\url{https://www.youtube.com/watch?v=j_3aYj5Hehk}} where robots are under control law $g^G$, and \href{https://www.youtube.com/watch?v=4X_gxu2PiWQ}{here}\footnote{\url{https://www.youtube.com/watch?v=4X_gxu2PiWQ}} in the scenario of Proposition~\ref{cor:individual-brake} where robot $3$ is the robot braking unexpectedly. 


\section{Conclusions and perspectives}
\label{sec-conclusion}

This paper turns the intuitive concept of priorities into a powerful mathematical tool to describe homotopy classes of solutions to the multirobot coordination problem. The first main result of this paper is that homotopy classes of feasible paths in the coordination space are uniquely encoded by priority graphs. Priorities are thus a meaningful unique representative of homotopy classes. Then, inspired by works on elastic strips~\cite{Quinlan1993,Brock2002} which revealed that controlling a robotic system under homotopic contraints can allow for more reactivity than executing a planned trajectory, we proposed to consider a priority graph as given, encoding an assigned homotopy class, and we designed a control scheme guaranteeing that the resulting trajectory of robots in the coordination space remains within the assigned homotopy class. Importantly, thanks to the freedom of action within the assigned homotopy class, the proposed control scheme allows for the deceleration or even stop of some or all robots for an arbitrary long time. That is a valuable property, in particular in the perspective of an implementation in an autonomous driving context where vehicles and pedestrians sharing the road results in a particularly unpredictable environment where many events may require some momentary deceleration to ensure safety of all road users. 

Three main perspectives of this work are particularly worth mentioning. First of all, the path-following assumption of Figure~\ref{fig-paths} is key to the definition of priorities and to the existence of homotopy classes. In real systems, perfect path following cannot be guaranteed as lateral control is based on imperfect mapping/localization data and imperfect actuators. Hence, future work should investigate which assumptions on lateral control still guarantee all the results of Section~\ref{sec:priority-encoded-homotopy-classes} which is the foundation of the priority-based approach. We believe that under guaranteed bounded uncertainty on lateral control, these results can be extended by considering the worst-case obstacle region considering all possible geometric paths. Secondly, this paper only focuses on the robust navigation in an assigned priority encoded homotopy class and the choice of a particular homotopy class -- which is key to time efficient coordination -- is not considered and should be investigated in future work. Note however that priorities can be obtained as a byproduct of all existing trajectory planning algorithms. One can simply use these planning algorithms and assign the priorities induced by the returned feasible path, yielding an equivalent efficiency. Finally, even though the control law $g^G$ effectively ensures remaining in the assigned homotopy class and is sufficient in applications which are not very sensitive to time efficiency, it should not be directly implemented in most scenarios. Alternatively, the interval $[\uL_i,g_i^G(s)]$ should be considered as the set of acceptable control values to apply for robot $i$ given the current state $s$ of the system. This control values interval can then be used as an input of a more complex controller performing some optimization. This approach proved efficient and is presented in the preliminary conference paper~\cite{Qian2015-MPC} where a model predictive control approach is used with a cost function aiming to respect a speed reference value and to penalize large control values, resulting in much smoother and efficient trajectories.





\bibliographystyle{elsarticle-num} 
\bibliography{biblio}

\end{document}